\documentclass[pdflatex,sn-mathphys-num]{sn-jnl}

\usepackage{mathpazo}
\usepackage{graphicx}%
\usepackage{multirow}%
\usepackage{amsmath,amssymb,amsfonts}%
\usepackage{amsthm}%
\usepackage{mathrsfs}%
\usepackage[title]{appendix}%
\usepackage{xcolor}%
\usepackage{textcomp}%
\usepackage{manyfoot}%
\usepackage{booktabs}%
\usepackage{algorithm}%
\usepackage{algorithmicx}%
\usepackage{algpseudocode}%
\usepackage{listings}%
\usepackage{tikz}%
\usetikzlibrary{positioning, fit}%
\usepackage{multirow}
\usepackage[utf8]{inputenc}
\usepackage{setspace}

\usepackage[dvipsnames]{xcolor}
\usepackage[most]{tcolorbox}

\definecolor{ourblue}{RGB}{30, 136, 229}

\theoremstyle{thmstyleone}%
\theoremstyle{thmstyletwo}%

\theoremstyle{thmstylethree}%

\begin{document}

\title[The Limits of Automatic Evaluation of Creativity in LLMs]{The Limits of Automatic Evaluation of Creativity in Large Language Models}


\author[1]{\fnm{Alessandro} \sur{Tutone}}\email{alessandro.tutone@studio.unibo.it}

\author[1]{\fnm{Giorgio} \sur{Franceschelli}}\email{giorgio.franceschelli@unibo.it}

\author*[2,1]{\fnm{Mirco} \sur{Musolesi}}\email{m.musolesi@ucl.ac.uk}

\affil[1]{\orgdiv{Department of Computer Science and Engineering}, \orgname{University of Bologna}, \orgaddress{\city{Bologna}, \country{Italy}}}

\affil[2]{\orgdiv{Centre for Artificial Intelligence, Department of Computer Science}, \orgname{University College London}, \orgaddress{\city{London}, \country{United Kingdom}}}


\abstract{
Large Language Models (LLMs) are increasingly capable of generating text that challenges human performance in domains requiring creativity, yet evaluating creativity in LLM-generated content remains a significant challenge. Here, we investigate whether current automatic evaluation methods can reliably capture human judgments of creativity. We collect human evaluations of human- and AI-generated short stories from the WritingPrompts dataset across 11 dimensions of creativity, and compare these judgments with automated objective metrics and LLM-as-a-Judge evaluations. Our experiments reveal substantial misalignment between automatic evaluations and human assessments. In particular, LLM-based judges exhibit a systematic preference for AI-generated stories, consistently favoring their stylistic characteristics over the unpredictability and other qualities of human-authored texts. Furthermore, correlation analyses show that widely used automatic metrics exhibit near-zero alignment with human judgments across both human- and AI-generated stories, suggesting that they fail to capture important dimensions of creativity. These findings highlight fundamental limitations in current approaches to the automatic evaluation of creative text and underscore the difficulty of reducing the multidimensional and subjective nature of creativity to computational metrics.}




\maketitle

\section*{Introduction}

In recent years, Artificial Intelligence (AI) models have become increasingly influential across a wide range of human activities, reshaping applications from academic research to industry. Following the public release of ChatGPT in 2022~\cite{openai2022introducing}, interest in generative AI~\cite{foster2019generative}, once largely confined to specialised research communities, has expanded rapidly. While initially focused primarily on text generation, this field has evolved to encompass models capable of synthesizing images~\cite{ramesh2021zero, ramesh2022hierarchical}, composing music~\cite{casini2025data}, and writing stories~\cite{yuan2022wordcraft, gomez2023confederacy}. As computational resources scale, the capabilities of these models continue to challenge human performance across increasingly diverse domains.

Among the many properties large language models (LLMs) and generative AI in general have been associated with, creativity is one of the most disputed. While previously considered the most human quality~\cite{bergson1912creative}, researchers are now increasingly debating the possibility that machines are not only potentially creative but also sometimes as creative as (or even more creative than) humans. From a philosophical perspective, depending on the assumed definition of creativity, it is indeed possible to reach contradictory conclusions. When only looking at the observable properties of the generated product, i.e., whether it is original and effective~\cite{standard_def_creativity} or novel, surprising, and valuable~\cite{creative_mind}, LLM outputs can arguably be as creative as humans'~\cite{wang2024ai}. However, when considering other, latent perspectives, e.g., those related to the process, the creator, and its environment~\cite{rhodes1961analysis}, the conclusions are the exact opposite~\cite{updating_std_def}. Because of this, analyses of artificial creativity should be more nuanced~\cite{franceschelli2026creativity}, and may consider multi-level scales rather than binary outcomes~\cite{oecd2}. The same apparently contradictory dichotomy can be observed in experimental findings as well, where, depending on the specific evaluation settings, it is possible to arrive at different results.

Indeed, how creativity can be evaluated in practice remains an open question. First, creativity depends on the task at hand: while various tests aim to assess general human divergent thinking abilities~\cite{guilford1967creativity, torrance1966torrance, olson2021naming}, they assume that such abilities correlate with creativity, an assumption that remains unproven not only for AI but also for humans~\cite{baer1993creativity}. Second, creativity is intrinsically subjective and depends on the observer's prior knowledge, skills, and preferences within a specific domain~\cite{katz1982subjective}, making standardized, ground-truth-based metrics for evaluating creativity particularly difficult to establish.

In particular, the evaluation methodologies adopted by prior research on creativity and LLMs are highly fragmented. Human evaluation, while ideal for capturing the subjective value, novelty, and surprise of a generated artifact, is difficult to collect, does not enable straightforward experimental validation and replication~\cite{belz2023non}, and depends on evaluators' skill level~\cite{davis2024chatpgt}. Therefore, in addition to or in place of human evaluation, the majority of studies adopt one or both of two different strategies: reliance on quantitative metrics~\cite{ismayilzada2025evaluating}, which are easy to compute and can capture several aspects of generated text but work at a lower level of abstraction; or adoption of the LLM-as-a-Judge paradigm~\cite{zheng2023judging}, which can better simulate human evaluation and allows for result reproduction, but assume LLMs to have evaluative capabilities similar to humans'.

\begin{figure}
    \centering
    \includegraphics[width=\linewidth]{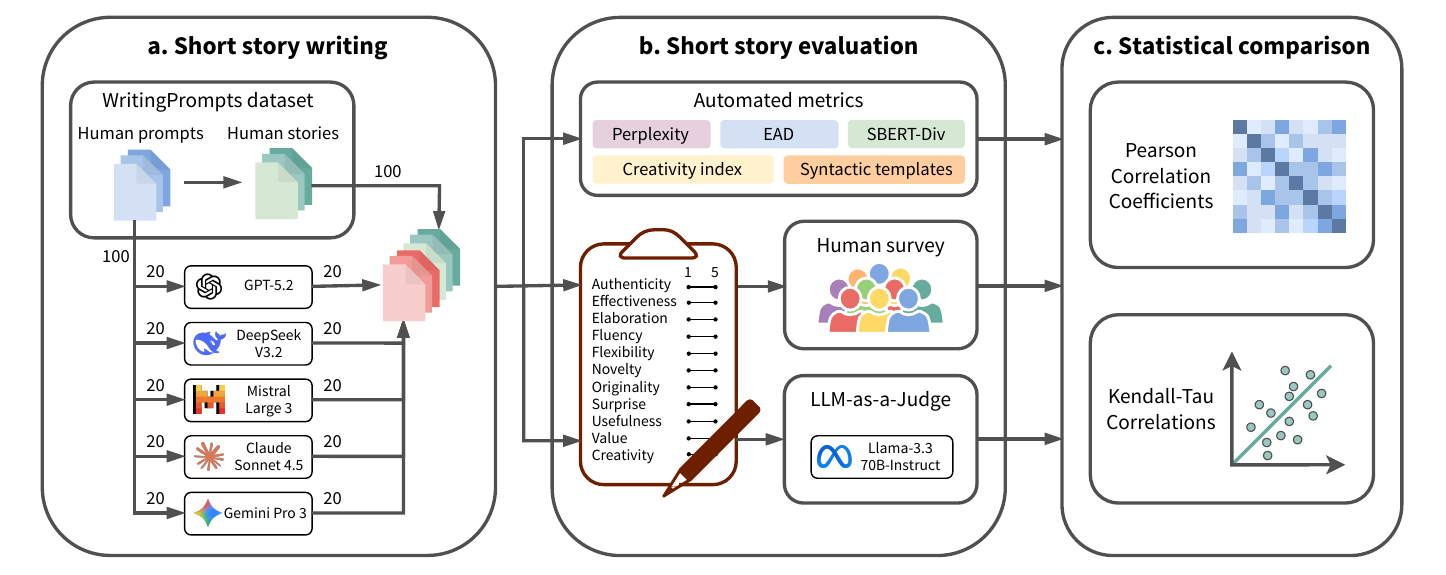}
    \caption{\textbf{Visual summary of our experimental pipeline.} \textbf{a. Short story writing}: We collected 100 human-written stories from WritingPrompts, and we complemented them with an additional 100 stories generated by five different state-of-the-art large language models. \textbf{b. Short story evaluation}: This heterogeneous dataset was subject to three different types of creativity evaluation: five automatically computed evaluation metrics, i..e, perplexity, expectation-adjusted distinct (EAD) $n$-grams, SentenceBERT embedding diversity (SBERT-Div), Creativity Index, and a series of syntactic-template-based scores; qualitative human survey on 11 dimensions of creativity; LLM-as-a-Judge ratings on the same 11 dimensions. \textbf{c. Statistical comparison}: We compare all these scores through appropriate correlation tests, which highlight the limitations of automatic evaluation schemes in assessing creativity.}
    \label{fig:intro_diagram}
\end{figure}

In this paper, we investigate whether current automatic evaluation schemes are appropriate for assessing human and artificial creativity, and the extent to which they correlate with human judgements of creativity and its multiple dimensions. Specifically, we curate and analyze a comparative dataset comprising 100 human-authored creative texts from the WritingPrompts dataset~\cite{fan2018hierarchical} and 100 LLM-generated texts. We then conduct an extensive survey to collect human ratings of their creativity and 10 related concepts, alongside ratings produced by an LLM-as-a-Judge framework for the same concepts and scores from five commonly used quantitative metrics. Figure~\ref{fig:intro_diagram} summarizes our experimental setting.

By comparing aggregate scores for human- and LLM-generated texts and analyzing correlations between different metrics in the context of story writing, we find that none of the automated evaluation methods, including LLM-as-a-Judge scores, provides a reliable proxy for human evaluation. The majority of automatically computed evaluation metrics show negative or no correlation with human scores, while only the less subjective creative dimension of elaboration exhibits a positive, albeit weak, correlation with LLM judgments. Analyses at a finer granularity suggest that this misalignment may arise because LLM judges primarily consider apparent, surface-level properties of the text rather than the broader spectrum of semantic dimensions considered by humans. Finally, the LLM-as-a-Judge paradigm exhibits a strong bias in favor of AI-generated stories, further limiting its suitability for automated creativity assessment and highlighting the need for alternative evaluation methods and strategies.

There already exist a few related works on the spurious, if not absent, correlation between human evaluations and automatic metrics~\cite{lu2026rethinking, saakyan2026death}, as well as the self-preference bias of LLM judges~\cite{panickssery2024llm, thakur2025judging}. We move one step further and systematically compare such metrics on both human and artificial artifacts, providing evidence of poor correlation for both, suggesting where and why these issues arise, and carefully discussing the implications of our findings for researchers in the field of LLMs.

\section*{Results}

\subsection*{Automatically computed evaluation metrics are not adequate to quantify creativity}
\begin{figure}[htbp]
    \centering
    \includegraphics[width=1\textwidth]{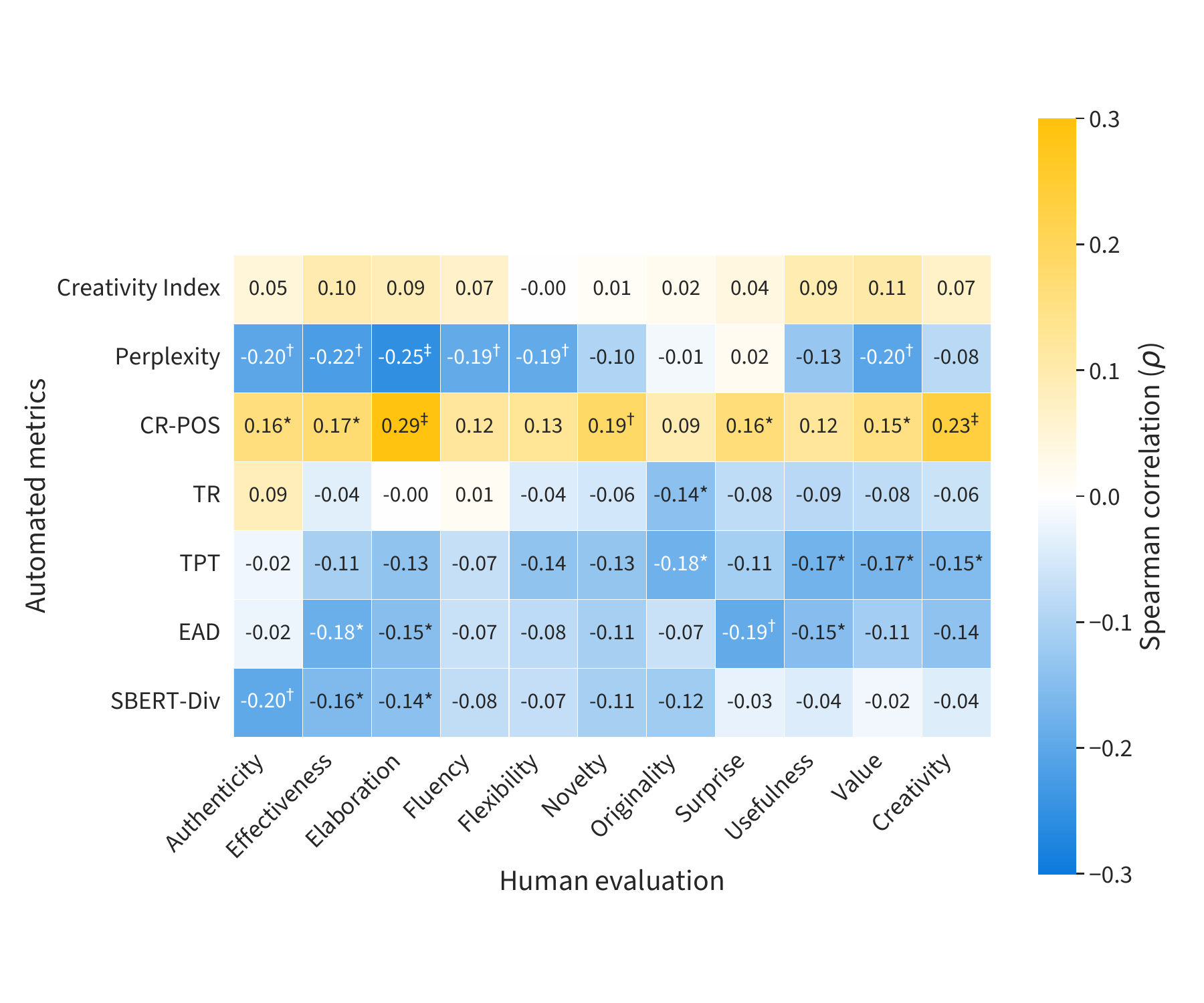 }
    \caption{\textbf{Correlation between automatically computed evaluation metrics and human evaluation.} Spearman correlation matrix between the seven automatically computed evaluation metrics and the eleven dimensions from human evaluation for both human-written and LLM-generated short stories. Statistical significance is denoted by single-character typography: * ($p < .05$), $\dagger$ ($p < .01$), and $\ddagger$ ($p < .001$).}
    \label{fig:correlation_heatmap}
\end{figure}

We selected seven of the most widely adopted automatically computed evaluation metrics for evaluating (dimensions of) creativity, namely, Creativity Index~\cite{lu2024ai}, perplexity~\cite{ismayilzada2025creative}, Expectation-Adjusted Distinct (EAD) $n$-grams~\cite{liu2022rethinking}, SentenceBERT embedding cosine diversity (SBERT-Div)~\cite{kirk2023understanding}, and three scores based on syntactic templates~\cite{shaib2024detection}, i.e., Compression Ratio of POS tags (CR-POS), Template Rate (TR), and Templates-Per-Token (TPT). We investigated the extent to which each metric correlates with subjective human evaluations of creativity in short stories by computing pairwise correlations. We used Spearman's rank correlation coefficient ($\rho$), which captures monotonic relationships between continuous automatically computed evaluation metrics and ordinal human Likert scales. Figure \ref{fig:correlation_heatmap} presents the resulting correlation heatmap.

As is apparent, automatically computed evaluation metrics show an overall lack of strong alignment with human evaluations. The vast majority of correlation coefficients fall within the $[-0.2,0.2]$ range, indicating very weak to negligible relationships. This suggests a substantial disconnect between automated text evaluation and human assessment of narrative creativity. Despite these generally weak correlations, several notable trends emerge.

Paradoxically, the metric explicitly designed to quantify creativity, i.e., Creativity Index, demonstrates almost no correlation with human-evaluated \textit{Creativity} ($\rho = 0.07$). Furthermore, it fails to exceed $\rho = 0.11$ across any of the other ten subjective dimensions. This finding highlights the difficulty of reducing a highly subjective and complex construct such as creativity to rigid mathematical and data-driven formulations of \textit{derivativeness}.

Perplexity yields similarly weak associations, exhibiting consistently weak negative correlations across almost all human-rated dimensions, with the strongest correlations for \textit{Effectiveness} ($\rho = -0.23$) and \textit{Elaboration} ($\rho = -0.21$). This indicates a slight tendency for humans to assign higher ratings to texts with \textit{lower} perplexity. From a human reader's perspective, highly unpredictable (high-perplexity) texts may be perceived as disjointed or less effective, whereas more familiar and predictable texts may be rewarded for their readability.

Analogous to Perplexity, which measures a model's internal uncertainty, SBERT-Div captures the degree of semantic divergence between sentences. Its overall negative correlation with human creativity evaluations suggests that greater semantic variation does not necessarily translate into higher perceived creativity and may instead be detrimental when distinct concepts are introduced without sufficient narrative connection. A creatively successful story therefore appears to require a balance between semantic diversity and thematic coherence, rather than maximizing either in isolation.

Furthermore, both TPT and EAD exhibit consistent, albeit weak, negative correlations with human evaluation metrics such as \textit{Surprise}, \textit{Originality}, and \textit{Value}. Although none of these correlations reach statistical significance, their consistent direction may nevertheless offer some suggestive insights into how these automatically computed evaluation metrics relate to human perceptions of creativity. For TPT, the negative correlation may indicate that texts characterized by greater structural regularity and denser use of repetitive templates tend to receive slightly lower ratings for creativity and surprise. Conversely, since higher EAD reflects greater lexical diversity (i.e., fewer repeated words), its negative correlation may suggest that greater lexical variation does not necessarily translate into higher perceived creativity. One possible explanation is that excessive lexical variation may reduce narrative or semantic cohesion, leading human evaluators to perceive it less as genuine surprise or originality and more as a lack of consistency in language use.

The strongest positive correlation in the matrix is observed between CR-POS and human-evaluated \textit{Elaboration} ($\rho = 0.29$). Although this is a weak association, it suggests that syntactic regularity may contribute to the perception of elaboration. Because higher CR-POS values indicate greater redundancy in syntactic structure, this result suggests that texts with more repetitive and structurally regular patterns may be perceived as more detailed or ``elaborated''. One possible explanation is that descriptive writing often employs parallel sentence structures and recurring syntactic patterns to organize and convey information. Thus, syntactic repetition may contribute to an impression of elaboration, even when it does not necessarily correspond to greater substantive content.

Overall, these correlations suggest that while specific syntactic features, such as POS variance, may capture certain aspects of human-perceived elaboration, automated statistical metrics have limited ability to serve as proxies for human aesthetic judgments of creativity.

\subsection*{LLMs are strongly biased in favor of AI-generated stories}
Having established the inadequacy of traditional statistical metrics, we next investigated whether large language models (LLMs) align with human judgments of creativity, and thus whether the ``LLM-as-a-Judge'' paradigm~\cite{zheng2023judging} can serve as a proxy for human creativity evaluation.

However, an initial analysis of the scores assigned by the LLM to human- and LLM-written stories revealed a substantial misalignment with human evaluations. As shown in Table \ref{tab:subjective_results} (left), human evaluators found the two sets of texts to be largely indistinguishable in terms of quality, with only \textit{Authenticity} and \textit{Elaboration} showing statistically significant differences in favor of LLM-generated stories. In stark contrast, the LLM-as-a-Judge evaluations exhibit a systematic self-preference bias. Compared with the human evaluations, the LLM scores differ significantly across all 11 dimensions, as detailed in Table \ref{tab:subjective_results} (right).

\begin{table}[ht]
    \centering
    \renewcommand{\arraystretch}{1.1} 
    \setlength{\tabcolsep}{4.6pt}
    \begin{tabular}{@{} l ccc c ccc @{}} 
        \toprule
         & \multicolumn{3}{c}{\textbf{Evaluated by Humans}} && \multicolumn{3}{c}{\textbf{Evaluated by LLM}} \\
        \cmidrule{2-4} \cmidrule{6-8}
        \multirow{-2.5}{*}{\textbf{Dimension}} & \textbf{Human Texts} & \textbf{AI Texts} & \textbf{\textit{p}-value} && \textbf{Human Texts} & \textbf{AI Texts} & \textbf{\textit{p}-value} \\
        \midrule
        Authenticity  & $3.27 \pm 0.07$ & $3.50 \pm 0.09$ & $\mathbf{.025}$ && $4.42 \pm 0.09$ & $4.97 \pm 0.03$ & $\mathbf{<.001}$ \\
        Effectiveness & $3.32 \pm 0.08$ & $3.48 \pm 0.09$ & $.239$          && $4.61 \pm 0.06$ & $5.00 \pm 0.00$ & $\mathbf{<.001}$ \\
        Elaboration   & $3.17 \pm 0.09$ & $3.54 \pm 0.09$ & $\mathbf{.005}$ && $4.86 \pm 0.04$ & $5.00 \pm 0.00$ & $\mathbf{<.001}$ \\
        Fluency       & $3.44 \pm 0.09$ & $3.63 \pm 0.09$ & $.191$          && $4.43 \pm 0.07$ & $5.00 \pm 0.00$ & $\mathbf{<.001}$ \\
        Flexibility   & $3.04 \pm 0.08$ & $3.23 \pm 0.08$ & $.130$          && $4.34 \pm 0.09$ & $4.98 \pm 0.01$ & $\mathbf{<.001}$ \\
        Novelty       & $3.03 \pm 0.09$ & $3.18 \pm 0.09$ & $.293$          && $4.20 \pm 0.11$ & $4.90 \pm 0.04$ & $\mathbf{<.001}$ \\
        Originality   & $3.13 \pm 0.09$ & $3.18 \pm 0.09$ & $.845$          && $4.68 \pm 0.06$ & $5.00 \pm 0.00$ & $\mathbf{<.001}$ \\
        Surprise      & $3.20 \pm 0.09$ & $3.07 \pm 0.09$ & $.271$          && $4.77 \pm 0.06$ & $4.90 \pm 0.05$ & $\mathbf{.013}$  \\
        Usefulness    & $3.03 \pm 0.08$ & $3.12 \pm 0.09$ & $.542$          && $3.53 \pm 0.17$ & $4.07 \pm 0.17$ & $\mathbf{.004}$  \\
        Value         & $2.90 \pm 0.09$ & $3.08 \pm 0.11$ & $.203$          && $4.65 \pm 0.06$ & $5.00 \pm 0.00$ & $\mathbf{<.001}$ \\
        Creativity    & $3.16 \pm 0.09$ & $3.27 \pm 0.09$ & $.569$          && $4.90 \pm 0.04$ & $5.00 \pm 0.00$ & $\mathbf{.004}$  \\
        \bottomrule
    \end{tabular}
    \caption{\textbf{Human and LLM scores on human-authored and LLM-written stories.} Comparison of the 11 dimensions of creativity as evaluated by human and LLM judges on stories authored by humans versus AI. Values are reported as mean $\pm$ standard error. Statistical significance (\textit{p}-values) of the differences between AI- and human-authored stories is derived from the Mann-Whitney U Test, and is reported in bold when significant ($p < .05$).}
    \label{tab:subjective_results}
\end{table}

Furthermore, the LLM exhibits a pronounced ceiling effect, assigning perfect scores to all LLM-generated texts across several dimensions, including \textit{Effectiveness}, \textit{Elaboration}, \textit{Fluency}, \textit{Originality}, \textit{Value}, and \textit{Creativity}. Thus, for these dimensions, the LLM assigns no variance to texts generated by LLMs, even when they originate from different models. In contrast, its evaluations of human-authored texts are both lower and more variable. This marked asymmetry suggests a strong preference for synthetic text and indicates that, in this setting, the LLM-as-a-Judge approach may systematically favor outputs that resemble its own learned linguistic and structural patterns.

Crucially, the entire evaluation process was strictly blind. Every text presented to both the human and LLM evaluators was completely devoid of source attribution or metadata; only the raw narrative text was evaluated. This ensures that the observed human variance and the LLM's systematic self-preference were not driven by explicit source information, but rather reflect differences in how the evaluators responded to the texts themselves.

\subsection*{LLMs are unreliable judges of human creativity as well} 
\begin{figure}[htbp]
    \centering
    \includegraphics[width=1\textwidth]{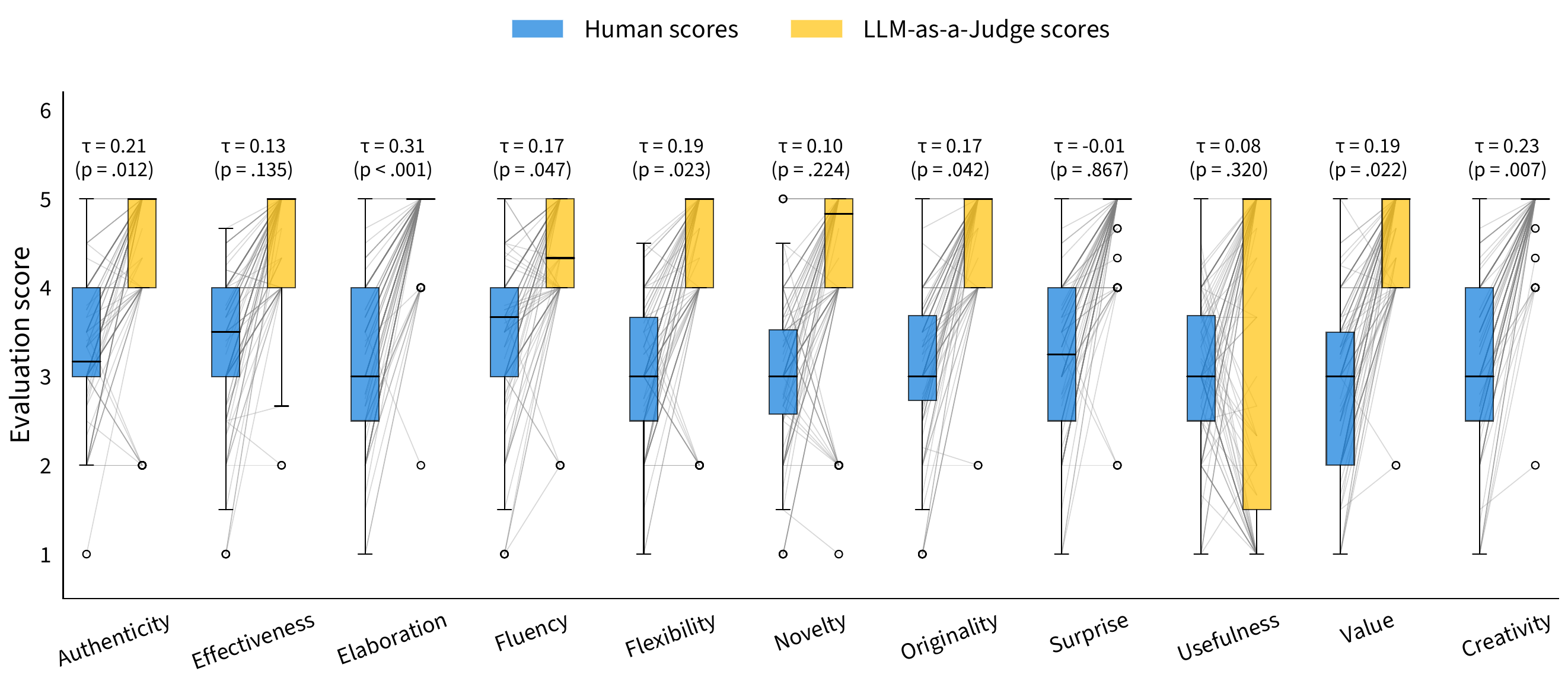}
    \caption{\textbf{Comparison between human evaluations and LLM judgments for human-written stories.} For each of the 11 subjective dimensions, we report Kendall's rank correlation coefficient ($\tau$) and its $p$-value (above), together with paired boxplots showing scores assigned to the same texts linked with straight lines.}
    \label{fig:paired_boxplot_human_texts}
\end{figure}

While the LLM-as-a-Judge paradigm proved inadequate for evaluating LLM-generated stories, it may still provide a useful automated approach for evaluating the creativity of human-written texts, given the similar variability of their scores. To quantify the alignment between machine evaluations and human judgment, we applied Kendall’s rank correlation coefficient ($\tau_b$), and we reported it together with a visual comparison of human and LLM scores for the same stories in Figure \ref{fig:paired_boxplot_human_texts}.

The results provide evidence of a severe misalignment between human evaluations and the LLM-as-a-Judge paradigm. Among the 11 subjective dimensions, only \textit{Elaboration} achieves a $\tau_b$ above 0.30, indicating that the judge exhibits at most a modest rank association with human evaluations.

For four of the evaluated dimensions (\textit{Effectiveness}, \textit{Novelty}, \textit{Surprise}, and \textit{Usefulness}), the correlations do not reach statistical significance ($p > 0.05$). This indicates that, for these dimensions, LLM-based evaluations provide little evidence of predictive validity with respect to human judgments. For instance, an LLM assigning a perfect 5.0 score to a story for \textit{Surprise} provides no statistically reliable basis for predicting how a human evaluator will score the same text on that dimension. Furthermore, \textit{Surprise} displays a near-zero correlation ($\tau_b = 0.01$, $p = 0.867$), providing strong evidence that the LLM's assessment of narrative unpredictability is fundamentally misaligned with the human experience of surprise.

Even among the seven statistically significant ($p < 0.05$) dimensions, the strength of the agreement remains remarkably weak. The highest correlation observed in the entire study occurs for \textit{Elaboration} ($\tau_b = 0.31$, $p < 0.001$). This dimension may be considered among the more objective of the subjective metrics, as it is closely related to measurable properties, such as text length and level of detail. The LLM therefore demonstrates some capacity to recognize structurally detailed texts; however, the substantially lower correlation for \textit{Creativity} ($\tau_b = 0.23$) suggests a marked limitation in its ability to capture the aesthetic or emotional qualities that make such details creatively effective.

\subsection*{Inner and outer LLM uncertainties are inconsistent} 
Perplexity is often used as a measure of uncertainty and surprise~\cite{basu2021mirostat}, as it represents the exponential average unexpectedness of each token under the model's distribution. We investigated whether this inner, implicit score correlates with the outer, explicit scores generated using the LLM-as-a-Judge paradigm, and thus whether there is consistency between inner and outer notions of uncertainty and unexpectedness. For this reason, we computed the perplexity under the same LLM used in the LLM-as-a-Judge paradigm and compared it with the predicted scores.

\begin{figure}[htbp]
    \centering
    \includegraphics[width=1\textwidth]{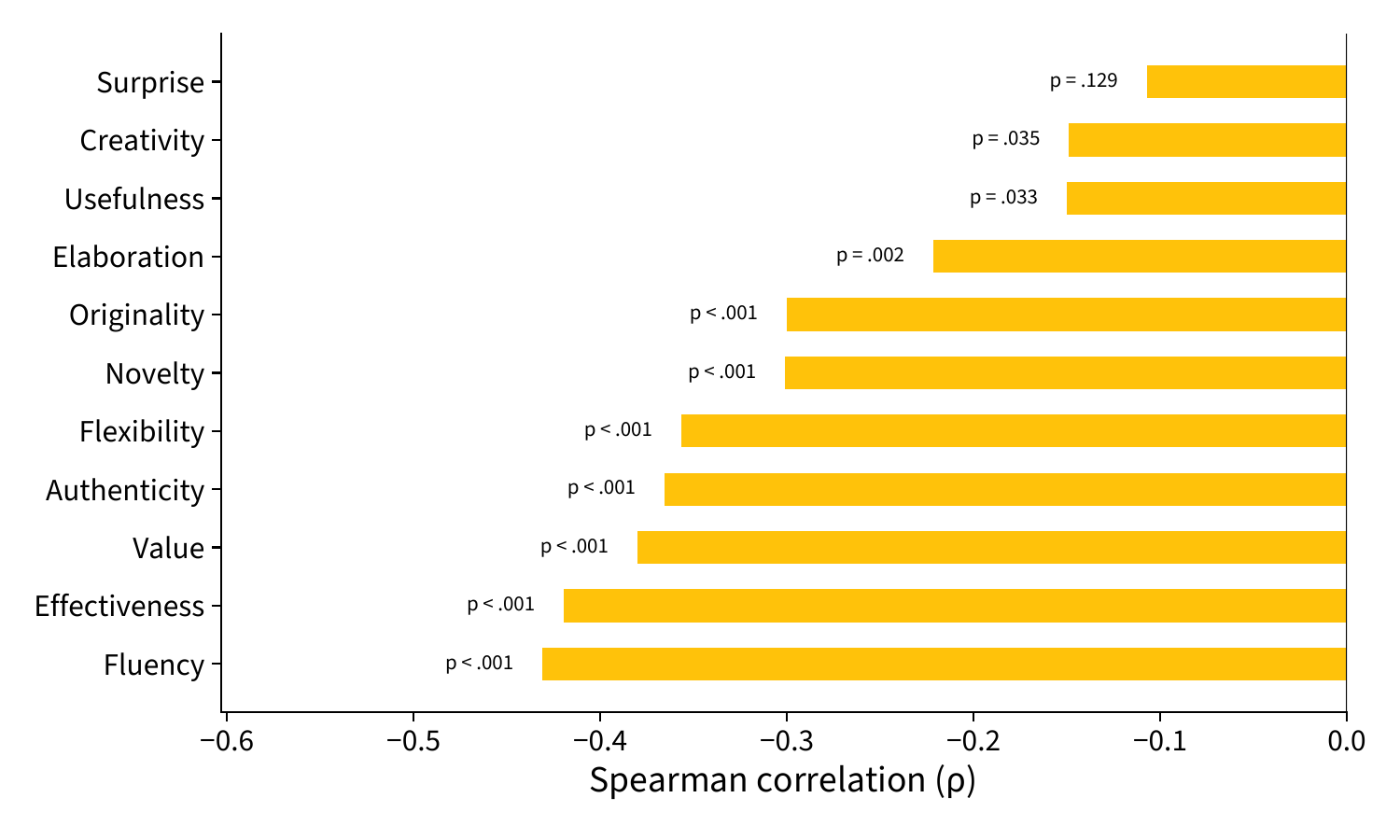}
    \caption{\textbf{Correlation between perplexity and LLM-as-a-Judge evaluations.} Spearman correlation between perplexity and the LLM-as-a-Judge evaluations computed under the same LLM. We report the $p$-values next to each bar. The chart only shows negative values on the $x$-axis since no subjective dimension has a positive correlation with perplexity.}
    \label{fig:perplexity_correlation_barchart}
\end{figure}

As reported in Figure \ref{fig:perplexity_correlation_barchart}, our results show that they are severely misaligned. In particular, perplexity has a negative correlation with all 11 subjective dimensions. While this is expected for more qualitative dimensions, such as fluency, effectiveness, and value, the weak negative correlation with surprise and creativity, together with the moderate negative correlation with originality and novelty, highlights a profound separation between mechanistic, token-based uncertainty and predicted, sentence-based unexpectedness.

\subsection*{The meaning of creativity differs between humans and LLMs} 
In the preceding sections, we have shown that automatic assessment differs substantially from human assessment of creativity, as the two are poorly correlated and sometimes measure entirely separate concepts. To better understand this conceptual misalignment, we finally investigated which dimensions contribute most to the overall notion of creativity in both human and LLM subjective evaluations.

\begin{figure}[htbp]
    \centering
    \includegraphics[width=1\textwidth]{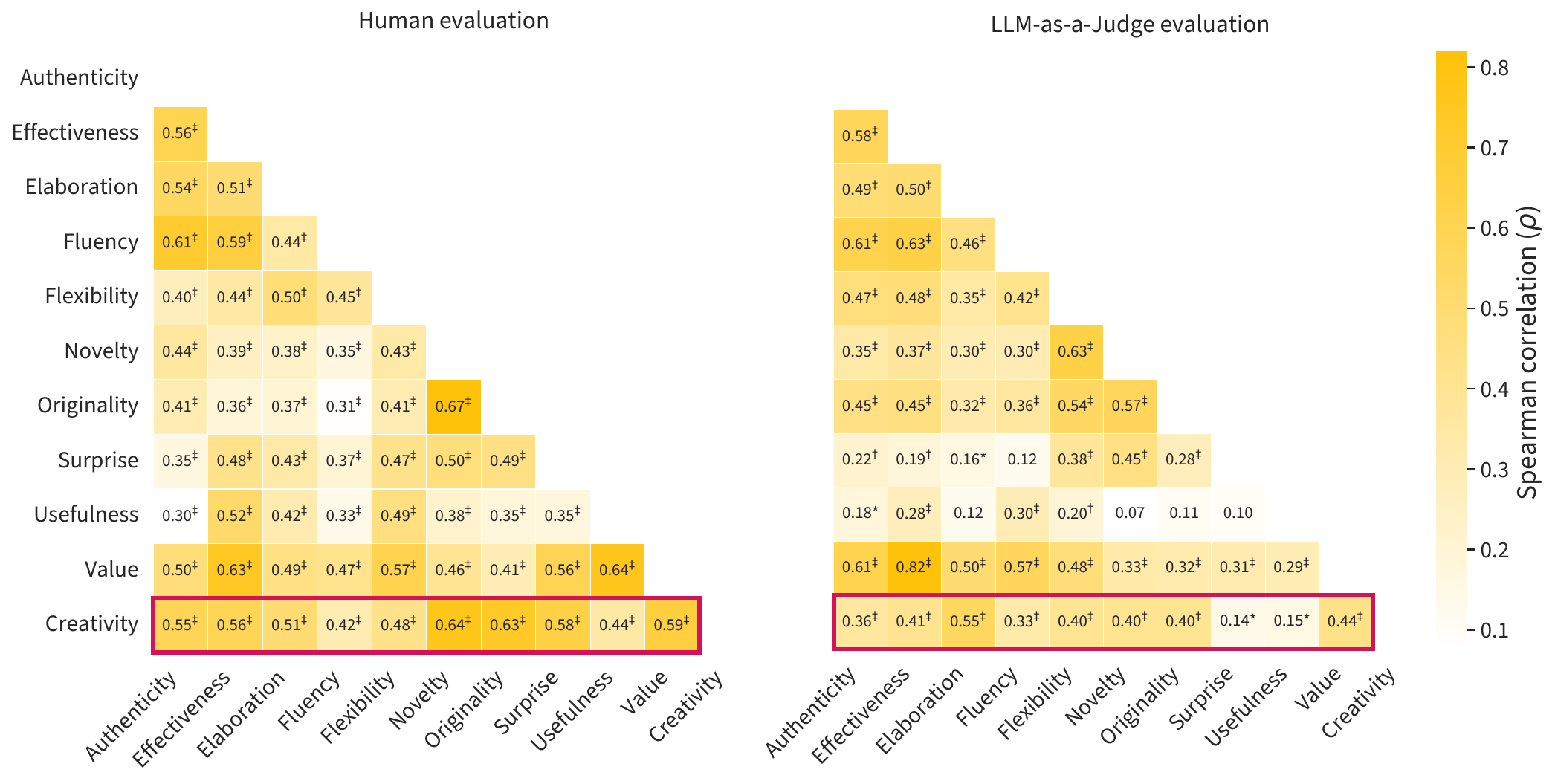}
    \caption{\textbf{Correlation of subjective dimensions evaluated by humans and LLM.} Spearman correlation between different subjective dimensions when evaluated by humans (left) and when evaluated by the LLM (right). Correlations with the overall \textit{Creativity} score are highlighted in red. Statistical significance is denoted by single-character typography: * ($p < .05$), $\dagger$ ($p < .01$), and $\ddagger$ ($p < .001$).}
    \label{fig:half_correlation_matrices}
\end{figure}

As reported in Figure \ref{fig:half_correlation_matrices}, human evaluation of \textit{Creativity} is linked to all other 10 dimensions, but shows stronger correlations with \textit{Novelty}, \textit{Originality}, \textit{Surprise}, \textit{Value}, \textit{Effectiveness}, and \textit{Authenticity}, i.e., the key dimensions in Boden's and Runco's definitions of creativity~\cite{creative_mind,updating_std_def}. In contrast, LLM evaluation of \textit{Creativity} correlates less strongly with the other dimensions and shows no correlation with \textit{Surprise} or \textit{Usefulness}. Moreover, its strongest correlation is with \textit{Elaboration}, highlighting a semantic misalignment between how humans and LLMs understand creativity that may help explain the inadequacy of automatic metrics for evaluating creativity. Notably, the highest correlation among LLM-evaluated dimensions is between \textit{Value} and \textit{Effectiveness}, which are also commonly connected in the creativity literature~\cite{standard_def_creativity}. However, \textit{Value} encompasses other qualities as well: while the LLM-as-a-Judge paradigm appears to capture more surface-level ones, such as \textit{Fluency}, it shows little association with more semantic dimensions, such as \textit{Usefulness}, which instead correlates with \textit{Value} in human evaluations. Similarly, \textit{Novelty} correlates with \textit{Originality}, but more strongly with \textit{Flexibility}. These patterns may again reflect a greater reliance of LLM evaluation on superficial textual properties rather than on deeper semantic aspects of creativity.

\section*{Discussion}


In this study, we carried out a comprehensive, multifaceted evaluation to quantify the relationship between human and artificial creativity assessment. Rather than isolating specific aspects of creativity, relying solely on subjective human ratings, or depending entirely on individual statistical proxies, our work adopted a holistic approach. We investigated the viability of automated methods, specifically the LLM-as-a-Judge paradigm and quantitative metrics such as the Creativity Index, using a balanced comparative dataset of 100 human-authored and 100 AI-generated short stories.

By combining these automated statistical metrics with an extensive baseline of subjective human judgments, we were able to systematically study the alignment between machine and human assessments. This comprehensive evaluation allowed us not only to measure these dimensions individually but also to characterize the relationships between quantitative metrics and human-perceived creativity. Ultimately, our results revealed a substantial disconnect: LLM judges exhibited a strong preference for machine-generated text and failed to reliably capture deeper semantic aspects of human creativity. Similarly, automatically computed evaluation metrics showed weak or no correlation with human and LLM judgments, highlighting their limitations in capturing complex and subjective dimensions such as narrative uncertainty and surprise.


More specifically, none of the widely adopted automatically computed evaluation metrics used to evaluate creativity and its closely related dimensions provided a sufficiently reliable proxy for human judgments. The overall misalignment between these metrics and human perception suggests a fundamental disconnect between quantitative text evaluation and human appreciation of narrative creativity. These findings demonstrate that current quantitative frameworks for measuring creativity remain fundamentally limited in their ability to capture the subjective and multifaceted nature of creativity. The mathematical formulations proposed over the years nevertheless represent valuable tools for analyzing stylistic and structural choices, as demonstrated by the statistically significant correlation between the compression ratio of POS tags (i.e., the degree of coherence and consistency in the use of syntactic templates throughout a text) and human evaluations of elaboration. Indeed, a recognizable and consistent writing style is a defining characteristic of the work of many great novelists~\cite{leech2007style}. However, using such automatically computed evaluation metrics as a replacement for human aesthetic judgment should be approached with caution, as they failed to capture the holistic and inherently complex nuances of human storytelling. Consequently, these findings underscore the need to critically re-examine the tasks for which these automated tools are deployed, suggesting that their use may be better suited to more objective, structurally focused contexts. Similarly, researchers should exercise caution when adopting automatically computed evaluation metrics to evaluate creative outputs and actively consider human-in-the-loop strategies.

On the other hand, the LLM-as-a-Judge paradigm exhibited two main issues. First, the LLM revealed a strong algorithmic bias in favor of machine-generated narratives. Second, its scores completely failed to correlate with human judgments regardless of the writer's origin. After further investigation, we found that this may stem from a significant misalignment regarding the meaning of creativity: while human scores of creativity correlated with its most prominent dimensions from the literature (i.e., novelty, value, surprise, originality, and effectiveness)~\cite{creative_mind,standard_def_creativity}, LLM scores of creativity correlated primarily with surface-level, syntactic properties such as elaboration. Notably, this misalignment is even more apparent when considering more profound, semantic properties such as usefulness, novelty, and surprise. In addition, we found a substantial disconnect between the extrinsic, predicted scores of surprise and originality and the internal properties of the LLM commonly associated with them, namely, its perplexity.

In general, this strong misalignment between human and LLM evaluations highlights several critical risks, especially when coupled with the tendency of LLMs to homogenize writing and reduce overall diversity over time~\cite{anderson2024homogeneization,doshi2024generative,kumar2025human,moon2025homogenizing}. Relying on LLMs to prepare training data, evaluate outputs, and fine-tune subsequent generations of models without human supervision may amplify existing biases. This practice can create a closed, self-reinforcing feedback loop in which models become increasingly reliant on evaluations that may not reliably distinguish the qualities of their algorithmic peers. Within the academic community, this poses a potentially serious systemic risk: AI-based peer-review systems and automated evaluators may favor LLM-generated or LLM-assisted research papers over those authored entirely by human researchers.
This issue is not confined to the creative domain; rather, it represents a broader systemic vulnerability. As described by Shumailov et al.~\cite{shumailov2023curse}, the recursive training of AI models on data generated by other AI models can lead to ``model collapse''. This phenomenon can introduce irreversible degradation in the resulting models, as successive generations progressively lose information about the underlying distribution of human-generated data. Consequently, models may lose the ability to reproduce rare concepts, outliers, and nuanced stylistic variations.

If commercial AI developers and academic institutions continue to rely on automated evaluators and synthetic data for non-objective tasks without grounding them in genuine human preferences, the field risks further homogenization of generative writing. This algorithmic echo chamber could produce models that are increasingly optimized for statistical regularity but less capable of capturing the aesthetic diversity of human writing, potentially influencing the stylistic preferences of human readers who rely on them. More broadly, it risks contributing to an algorithmic definition of creativity: one that treats the distinctive imperfections of human narratives as undesirable and favors the stylistic uniformity of machine-generated text.

\subsection*{Limitations}
While this study provides critical insights into the relationship between artificial intelligence and creative writing and the challenges of automating creativity evaluation, a few limitations must be acknowledged. First, our analysis was restricted to a single task and a single dataset, specifically, the WritingPrompts collection. The corpus was limited to 200 texts, a necessary constraint due to the bottleneck of human-in-the-loop evaluation; analyzing a larger subset of the hundreds of thousands of available prompts would yield more statistically robust results.

Second, our experimental design relied on a single large open-source LLM to act as the automated judge, a decision necessitated by computational resource constraints. Consequently, we could not examine whether evaluative biases vary across different proprietary AI models (e.g., GPT-5 or Claude 4), which may be subject to self-preference biases to different degrees.
Finally, the AI-generated narratives evaluated in this study were produced using the default generative parameters adopted by the respective vendors. We did not explore variations in temperature, sampling schemes (such as top-$p$~\cite{holtzman2020curious} or top-$k$~\cite{fan2018hierarchical}), or advanced prompt-engineering techniques, nor did we isolate differences in writing and evaluative capabilities between base foundation models and their instruction-tuned counterparts.

These limitations present clear avenues for future research. Subsequent studies should expand this methodology across diverse text domains, such as scientific writing, technical documentation, and journalistic reporting, to determine whether the observed algorithmic biases and structural dependencies persist beyond purely creative contexts. Future experiments should also investigate how altering generation parameters affects both the production and evaluation of narrative texts. Most importantly, future work should move beyond merely identifying these limitations toward actively developing new evaluative frameworks. Building on the findings of this study, researchers should aim to develop multidimensional and potentially multimodal evaluation metrics that account for algorithmic biases and better capture the nuanced realities of human reading.

\subsection*{Conclusions}

Our research provides insights into the relationship between human and machine creativity assessment, particularly into whether machines can accurately evaluate the creativity of short narratives through automatic quantitative metrics or the LLM-as-a-Judge paradigm.
Our findings revealed that automatically computed evaluation metrics are inadequate proxies for evaluating human creativity. They showed near-zero correlations with any of the 11 subjective dimensions evaluated by human judges, demonstrating that rigid, token-based approaches largely fail to capture the semantic and emotional qualities that human readers value. LLMs proved to be misaligned evaluators as well. The pronounced presence of a self-preference bias, together with a lack of correlation with the most subjective human scores, makes the LLM-as-a-Judge paradigm fundamentally unreliable for autonomous, human-free creativity assessment.
On a larger scale, this creates the risk of a closed, self-reinforcing feedback loop in which models fail to objectively assess their peers and penalize authentic human expression in favor of a highly predictable, low-variance production standard. The nuances and distinctive imperfections of human creations are profoundly complex and difficult to articulate, even by humans themselves; therefore, mathematically encoding them remains extremely challenging.
Until artificial systems can truly capture the richness of the human experience they attempt to describe, creativity may remain a deeply complex aspect of human cognition that current algorithms have yet to fully ``learn''.

\section*{Methods} \label{methods}

This study adopts a multi-stage experimental design. The pipeline is structured to enable methodological triangulation by contrasting statistical probability measures with human cognitive judgments. Establishing correlations between these measures is crucial for assessing the reliability of automated creativity evaluation.

\subsection*{Dataset composition}
This study uses a balanced dataset of 200 stories, comprising 100 randomly selected human-generated and 100 machine-generated stories. Both sets consist of texts paired with a unique writing prompt, which serves as the creative premise upon which the fictional short story is based. 
First, we populated the human-generated story partition by randomly sampling 100 prompt–story pairs from the WritingPrompts dataset~\cite{fan2018hierarchical}. The dataset comprises a large collection of user-submitted prompts sourced from Reddit’s WritingPrompts forum, each paired with a user-generated story responding to the corresponding prompt. Then, we populated the machine-generated story partition by randomly sampling a separate set of 100 prompts from the WritingPrompts dataset and providing them as inputs to LLMs. More specifically, we employed a diverse set of state-of-the-art models to ensure broad representation: \texttt{GPT-5.2}~\cite{gpt_5.2}, \texttt{DeepSeek-V3.2}~\cite{deepseekai2025deepseekv32}, \texttt{Mistral Large 3}~\cite{mistral3}, \texttt{Claude Sonnet 4.5}~\cite{claudesonnet45}, and \texttt{Gemini 3 Pro}~\cite{gemini3pro}. Each model was tasked with generating 20 stories, yielding a total of 100 machine-generated stories. The specific generation prompts are provided in Supplementary Information \ref{app:prompts_gen}.

While we preserved the source information for \textit{a posteriori} analytical purposes, all evaluations were conducted blindly, without identifiers distinguishing human-generated from AI-generated stories, to ensure fair comparisons and minimize potential biases.

\subsection*{Automatically computed evaluation metrics}

To quantify the statistical and structural properties of the considered texts, this study employs five distinct metrics commonly used to automatically assess creativity or one of its main dimensions: Creativity Index, Perplexity, Syntactic Templates, Expectation-Adjusted Distinct $n$-grams, and Sentence-BERT Embedding Diversity. Each metric evaluates the text at a specific granularity, from phrase-level lexical choices to high-level narrative structures.

\subsubsection*{Creativity Index}

To assess phrase-level originality, we utilize the \textit{Creativity Index} (CI) proposed by Lu et al.~\cite{lu2024ai}. This metric measures the extent to which a given text can be attributed to existing content on the web, thereby quantifying the ``derivative'' nature of the writing.

The core component of this metric is the concept of \textit{$L$-uniqueness}. Formally, for a given length $L$, the $L$-uniqueness of a text $\mathbf{x}$ is defined as the proportion of words $w \in \mathbf{x}$ that do not belong to any $n$-gram (with $n \geq L$) found in the reference corpus $C$:

\begin{equation}
    \mathcal{U}_L(\mathbf{x}) = \frac{|\{w \in \mathbf{x} \mid \forall g \in \text{ngrams}(\mathbf{x}): w \in g \land |g| \geq L \implies g \notin C \}|}{|\mathbf{x}|}
\end{equation}
Intuitively, a higher $L$-uniqueness value indicates greater novelty at the corresponding $L$-gram scale. The total CI is calculated by aggregating these uniqueness scores across a range of $n$-gram lengths, specifically within the bounds $[a, b]$:
\begin{equation}
    \text{CI}(\mathbf{x}) = \sum_{L=a}^{b} \mathcal{U}_L(\mathbf{x})
\end{equation}
In our implementation, we set the bounds $a=5$ and $b=12$ and use the RedPajama dataset~\cite{weber2024redpajama} as the reference corpus $C$, aligning with the standard configuration established in the original implementation of the metric~\cite{lu2024ai}.
To perform efficient $n$-gram matching between our target texts and this massive corpus (approximately 1.4 trillion tokens), we employ the Infini-gram engine~\cite{liu2024infini}. The entire process is orchestrated by the DJ Search algorithm~\cite{lu2024ai}, which efficiently retrieves the verbatim $n$-gram matches required to compute the $L$-uniqueness scores and, ultimately, the final CI.

\subsubsection*{Perplexity}

To target the ``surprise'' factor of a text, we employ \textit{Perplexity} (PPL), a score that represents the uncertainty in the predictions of a language model~\cite{lu2026rethinking}. Formally, it quantifies the branching factor of the model, i.e., the number of equally probable tokens that could follow the current context. Lower perplexity indicates that the model is more certain about the next token in a sequence, effectively narrowing down the possibilities. However, it is crucial to note that statistical confidence does not imply correctness; a model can exhibit low perplexity (high certainty) while still generating factually incorrect or hallucinated content~\cite{understanding_perplexity}.

Mathematically, given a probability model $P$ and a sequence of $N$ tokens $\mathbf{x} = (t_1, t_2, \dots, t_N)$, perplexity is defined as the exponential of the average negative log-likelihood:
\begin{equation}
    \text{PPL}(\mathbf{x}) = \exp\left( -\frac{1}{N} \sum_{i=1}^{N} \ln P(t_i \mid t_1, \dots, t_{i-1}) \right).
\end{equation}
A perplexity of 1 indicates that the model assigns a probability of 1.0 to all target tokens (i.e., the model has zero entropy). Conversely, a value higher than 1 reflects a higher degree of uncertainty. Intuitively, a perplexity of $k$ indicates that the model is as uncertain as if it were choosing uniformly among $k$ equally likely options. Thus, a very high value means that the model is uncertain about which token(s) to expect or was expecting something completely different.


In the context of creativity studies, perplexity can be seen as a proxy for statistical conventionality~\cite{lu2026rethinking}. Lower perplexity indicates text that follows conventional statistical patterns and is therefore more predictable, while higher perplexity suggests greater unpredictability, a trait often associated with human stylistic creativity~\cite{mitchell2023detectgpt}.

To compute this metric, our experimental pipeline utilizes the \texttt{Meta Llama-3-8B} (Base) model. Unlike older architectures such as \texttt{GPT-2}, \texttt{Llama-3}~\cite{dubey2024llama} operates with a significantly larger vocabulary (128k tokens) and a deeper semantic representation. This provides a more accurate measure of linguistic plausibility for modern generated text, ensuring a consistent statistical baseline for comparing state-of-the-art LLMs against human authors.

\subsubsection*{Syntactic Templates}

To evaluate the creativity of a text beyond lexical word choice, we employ \textit{Syntactic Templates}. By abstracting text into sequences of Part-of-Speech (POS) tags (e.g., \texttt{[DET NOUN VERB DET ADJ NOUN]}), we can evaluate the structural diversity of the narrative, i.e., \textit{how} it is constructed rather than \textit{what} is said.

The extraction of syntactic templates follows the methodology proposed by Shaib et al.~\cite{shaib2024detection}. The core concept relies on identifying abstract patterns that frequently repeat within a corpus. In particular, a template is defined as follows:
\begin{quote}
\textit{Given a sequence of tokens $T=(t_1,t_2,...,t_n)$ and a function $f$ that computes an abstraction over $T$ (e.g., part-of-speech tags), we define a \textbf{template} as a subsequence of abstractions over the tokens $f(T)$ that repeats at least $\tau$ times in $T$.}
\end{quote}

In our implementation, we utilize the Natural Language Toolkit (NLTK)~\cite{bird2009natural} to perform initial POS tagging, mapping each token to its corresponding syntactic role. We then employ the \texttt{diversity} library~\cite{shaib2024standardizing} to extract recurring patterns and compute the diversity metrics.

Once the templates are extracted, we employ three distinct metrics to quantify structural repetition:

\begin{itemize}
    \item \textbf{Compression Ratio of POS Tags (CR-POS)}. It quantifies the $n$-gram diversity of the syntactic structure using a lossless compression algorithm (\textit{e.g.}, gzip).
    The underlying principle is that compression algorithms are optimized to detect and condense repeated sequences; therefore, highly repetitive structures will compress to a much smaller size than diverse ones. Shaib et al.~\cite{shaib2024standardizing} demonstrated that this ratio effectively captures lexical and structural $n$-gram repetition.

    CR-POS is computed over the sequence $f(T)$ of all POS tags within a single text $T$. The ratio is defined as:
    \begin{equation}
        \text{CR-POS}(T) = \frac{|f(T)|}{|g(f(T))|}
    \end{equation}
  where $| \cdot |$ denotes the size in bytes and $g(\cdot)$ is a lossless compression function.
    
    A higher CR-POS value implies that the text contains greater redundancy in its syntactic structure (lower diversity), as the algorithm was able to compress it more significantly. Conversely, a lower value indicates a more structurally varied narrative.

    \item \textbf{Template Rate (TR)}. To evaluate structural redundancy at the individual narrative level, we adapt the template extraction methodology to compute a token-level coverage rate. For each text, we extract the most frequently recurring syntactic $n$-grams (``common templates''). We then calculate the proportion of the text's total tokens that belong to at least one frequent syntactic pattern~\cite{shaib2024detection}.

 TR aims to quantify how frequently templates appear across the whole narrative. Mathematically, for a given text $T$ consisting of $N$ tokens, let $M$ denote the set of tokens that are part of at least one ``common template''. The TR is defined as:
    \begin{equation}
        \text{TR}(T) = \frac{|M|}{N}
    \end{equation}
    A high TR suggests that the model (or author) frequently falls back on standard, repetitive grammatical structures rather than constructing novel sentences.
    
    \item \textbf{Templates-Per-Token (TPT)}. In practice, one text can contain multiple templates, and longer texts tend to contain more templates. To account for differences in text length at the individual level, TPT measures the density of structural repetition within a single narrative~\cite{shaib2024detection}.
    
    Similar to TR, we first extract the ``common templates'' for the specific text; then, we quantify the intensity of repetition by accounting for overlapping patterns. For each token in the text, we count the number of recurring template instances it participates in. Because templates are extracted using a sliding window, a single token might belong to multiple overlapping template matches. We sum these participation counts across all tokens and normalize by the total text length.
    
    Mathematically, for a text $T$ of $N$ tokens, let $k_j$ denote the number of identified template instances that encompass token $j$.

The TPT is defined as:
    \begin{equation}
        \text{TPT}(T) = \frac{\sum_{j=1}^{N} k_j}{N}
    \end{equation}
   A higher TPT indicates a denser concentration of repetitive structures per unit of text, meaning that tokens are more frequently embedded within multiple predictable syntactic patterns.
\end{itemize}

\subsubsection*{Expectation-Adjusted Distinct $N$-Grams (EAD)}
Evaluating the true diversity of generated text requires analyzing both surface-level variation and semantic novelty. The standard Distinct score~\cite{li2016diversity} is widely used to measure lexical diversity and is defined as $N/C$, where $N$ is the number of distinct tokens and $C$ is the total number of tokens.
%
However, it is also known to be negatively correlated with text length, as longer texts naturally contain more repeated words. To correct for this bias, the \textit{Expectation-Adjusted Distinct $n$-grams (EAD)} score~\cite{liu2022rethinking} has been proposed. This metric normalizes the observed count of unique tokens $N$ by its mathematical expectation.
EAD is calculated as:
\begin{equation}
    \text{EAD} = \frac{N}{\mathbb{E}[N]} \qquad \text{where} \quad \mathbb{E}[N] = V \left[ 1 - \left(\frac{V-1}{V}\right)^C \right].
\end{equation}
Here, $V$ is the total vocabulary size, and $C$ is the length of the specific text being evaluated. By comparing the observed diversity with the expected diversity of a random sampling process over a uniform distribution of size $V$, EAD provides a robust, length-independent measure of lexical richness.

A score close to 0 indicates that the text exhibits high lexical redundancy, relying on a restricted subset of the available vocabulary. Conversely, a score approaching 1 implies that the text exhibits a high degree of lexical diversity relative to its length, effectively utilizing a broad and varied distribution of tokens rather than converging on high-frequency patterns.

\subsubsection*{Sentence BERT Embeddings Diversity (SBERT-Div)}

To move beyond token-based matching and evaluate the conceptual diversity of the text, we utilize \textit{Sentence BERT} embeddings diversity (SBERT-Div)~\cite{kirk2023understanding}. We measure the semantic diversity of a text by calculating the complement of the pairwise cosine similarity between the sentences composing the narrative.

For a given text $T$ consisting of $K$ sentences $S = {s_1, ..., s_K}$, we first compute the dense vector embedding $u_i = \text{SBERT}(s_i)$ for each sentence using a pre-trained Sentence-BERT model. In particular, we employ the pre-trained \texttt{all-MiniLM-L6-v2} model~\cite{wang2020minilm}. While early SBERT implementations relied on fine-tuned \texttt{bert-base} architectures~\cite{reimers2019sentence}, recent benchmarks demonstrate that the \texttt{all-MiniLM} family offers a superior balance between computational efficiency and semantic accuracy~\cite{muennighoff2022mteb}.


Then, we calculate the average semantic similarity $\mu_{sim}$ across all unique sentence pairs to determine how semantically repetitive the text is:
\begin{equation}
    \mu_{sim} = \frac{2}{K(K-1)} \sum_{1 \le i < j \le K} \text{sim}(u_i, u_j)
\end{equation}
where the cosine similarity between two vectors is defined as:
\begin{equation}
    \text{sim}(u_i, u_j) = \frac{u_i \cdot u_j}{\|u_i\| \|u_j\|}
\end{equation}
Finally, the SBERT-Div metric is defined as the complement of this average similarity:
\begin{equation}
    \text{SBERT-Div} = 1 - \mu_{sim}
\end{equation}
A score close to 0 indicates a highly repetitive narrative, where sentences merely reiterate the same semantic concepts without advancing the plot, while a score close to 1 indicates a highly disjointed sequence of sentences that lack thematic connections, leading to an incoherent story. Therefore, a creatively successful narrative is expected to fall within an intermediate range, striking a delicate balance between narrative diversity and thematic coherence.

\subsection*{Subjective metrics}

While quantitative metrics provide reproducible statistical insights, creativity is inherently a subjective phenomenon that relies on human perception. To capture this nuance, we conduct a dual-track subjective evaluation involving both human annotators and an LLM-as-a-Judge approach. Crucially, both evaluations were conducted using a blind protocol: neither the human participants nor the LLM judge was provided with labels distinguishing human-authored texts from AI-generated ones.

To avoid the ambiguity of a single ``Creativity'' score, the evaluation has been decomposed into 11 distinct dimensions. These dimensions were strategically selected to cover a broad spectrum of theoretical definitions of creativity. Specifically, the metrics are grounded in established creativity literature, explicitly acknowledging Boden's triad of \textit{novelty}, \textit{surprise}, and \textit{value}~\cite{creative_mind}, as well as Runco's bipartite standard definition of \textit{originality} and \textit{effectiveness}~\cite{standard_def_creativity}. Additional dimensions were included to capture other prominent aspects of creativity, i.e., \textit{elaboration}, \textit{fluency}, and \textit{flexibility}~\cite{guilford1967creativity}; \textit{authenticity}~\cite{kharkhurin2014creativity}; and \textit{usefulness}~\cite{stein1953creativity}. Finally, we also considered an overall score of \textit{creativity}, which is useful for studying which dimensions correlate most strongly with the general subjective perception of creativity. Each dimension is rated on a 5-point Likert scale (1 = Poor, 5 = Excellent) and is described as follows:

\begin{enumerate}[itemsep=1pt]
    \item \textbf{Authenticity:} The perception of a genuine and personal voice. Does the text seem written with its own heartfelt style, or does it appear mechanical, artificial, or copied?
    
    \item \textbf{Effectiveness:} The ability to achieve the narrative goal. Does the story entertain and satisfy the prompt in a complete and convincing manner?
    
    \item \textbf{Elaboration:} The richness of details and depth of development. Does the text expand the main idea with vivid descriptions and complexity, rather than remaining superficial?
    
    \item \textbf{Fluency:} The linguistic smoothness and ease of idea generation. Does the text read naturally without grammatical hiccups, with ideas flowing logically from one to another?
    
    \item \textbf{Flexibility:} The variety of perspectives or themes. Does the text manage to connect different concepts, change viewpoints, or adapt to narrative constraints agilely?
    
    \item \textbf{Novelty:} The degree of conceptual innovation. Is the core idea fresh and new, or is it a rehash of well-known and overused concepts?
    
    \item \textbf{Originality:} The statistical rarity of the approach. Compared to other texts on the same topic, does this approach avoid clichés and commonplaces?
    
    \item \textbf{Surprise:} The ability to astonish the reader. Does the content take an unexpected turn, introduce a plot twist, or break common expectations regarding the topic?
    
    \item \textbf{Usefulness:} The relevance and applicability to the context. Is the text consistent with the initial request and does it function as a valid response to the prompt?
    
    \item \textbf{Value:} The intrinsic merit of the text. Is the reading experience enriching, interesting, or emotionally engaging? Does it possess a perceivable literary quality?
    
    \item \textbf{Creativity (Overall):} A holistic judgment on ingenuity and imagination. Has the author combined ideas uniquely to create a ``wow factor'' that distinguishes an inspired text from a tedious one?
\end{enumerate}

\subsubsection*{LLM-as-a-Judge evaluation}

To assess the feasibility of automating subjective creativity evaluation, we employ the LLM-as-a-Judge approach~\cite{zheng2023judging}, utilizing \texttt{meta-llama/Llama-3.3-70B-Instruct} as the evaluator. While proprietary models such as \texttt{GPT-5.2}~\cite{gpt_5.2} or \texttt{Claude Sonnet 4.5}~\cite{claudesonnet45} often exhibit marginally higher correlations with human judgments in creative tasks~\cite{zheng2023judging}, we select \texttt{Llama 3.3} because it represents a state-of-the-art open-weight model, ensuring a high-quality evaluation pipeline that is both reproducible and transparent within standard computational limits~\cite{dubey2024llama}. The model was tasked with acting as an objective literary critic, grading the texts according to the 11 definitions provided above.

A critical challenge in LLM-based evaluation is \textit{context contamination} or the ``Halo Effect'', where a model's judgment on one metric (e.g., \textit{Fluency}) inadvertently influences its scoring of subsequent metrics (e.g., \textit{Originality}) within the same context window.
Initially, we experimented with a \textit{batched prompting strategy}, instructing the model to output scores for all 11 metrics in a single inference pass. However, preliminary tests suggested that this approach led to high inter-metric correlations, failing to capture the nuances between well-written but derivative text (e.g., high \textit{Fluency}, low \textit{Originality}) and rough but creative text (e.g., low \textit{Fluency}, high \textit{Originality}).
To mitigate this bias, we adopt an \textit{isolated evaluation strategy}~\cite{liu2023g} and decompose the evaluation into independent inference calls. For every text-metric pair, the model is initialized with a fresh context containing only the system prompt (i.e., the role definition), the definition of the single metric being evaluated, and the target text.


This ensures that the score for dimension $a$ is calculated independently of the score for dimension $b$. In addition, to achieve deterministic and reproducible outputs, we set the generation temperature to 0.2 and the top-p parameter to 0.95, and we repeated the entire process three times. The template used for these atomic evaluations is provided in Supplementary Information \ref{app:prompts_judge}. By enforcing this separation, we aim to maximize the distinctiveness of each creativity dimension and reduce the noise introduced by the model's internal biases.

\subsubsection*{Human evaluation protocol}
To establish a ``ground truth'' for creativity, we deployed a custom web interface to collect subjective judgments. To ensure a broad representation of perspectives, the platform was distributed to a heterogeneous group of evaluators. The participants ranged from domain experts (e.g., professors and researchers) and university students across various faculties to laypeople outside the academic sphere (e.g., individuals from diverse professional backgrounds). This sampling strategy was intentionally designed to include individuals with varying levels of AI experience, ensuring that the evaluation reflects not only technical assessments but also the general public's perception of creativity.


Each participant was presented with one randomly selected story at a time and asked to rate it according to a structured rubric. To facilitate a broader evaluation, all narratives were made available in both English and Italian. For the AI-generated stories, the respective models were first prompted to generate the English version and subsequently to provide the Italian translation. Conversely, the human-authored stories were translated into Italian using \texttt{Google Gemini 3 Pro}~\cite{gemini3pro}.

Upon completion of an evaluation, responses were anonymously recorded in a cloud-based database. Alongside the scores, each entry automatically logged the following session metadata: \textit{timestamp}, \textit{session ID}, \textit{text ID}, \textit{language}, and \textit{authorship source} (Human or LLM). Additionally, to enable demographic analysis, we collected and recorded participants' \textit{native language}, \textit{gender}, \textit{age}, \textit{education level}, and \textit{self-reported AI expertise} (aggregated information about our pool of human evaluators can be found in Supplementary Information \ref{app:human_survey}). Participants retained full autonomy to evaluate multiple texts or conclude the session at any point, ensuring a voluntary and non-coercive testing environment.

\section*{Data availability}
All the generated and collected data used in this article are publicly available at: \url{https://github.com/pante31/LLM_Creativity.git}. The original prompts and human-written short stories have been taken from the WritingPrompts dataset \cite{fan2018hierarchical}, which is publicly available at: \url{https://huggingface.co/datasets/euclaise/writingprompts}.

\section*{Code availability}
To facilitate reproducibility and encourage further research, the complete source code and evaluation scripts are publicly available at: \url{https://github.com/pante31/LLM_Creativity.git}.

\newpage
\begin{appendices}

\section{Extended correlation matrices} \label{app:full_results}
\begin{figure}[htbp]
    \centering
    \includegraphics[width=1.\textwidth]{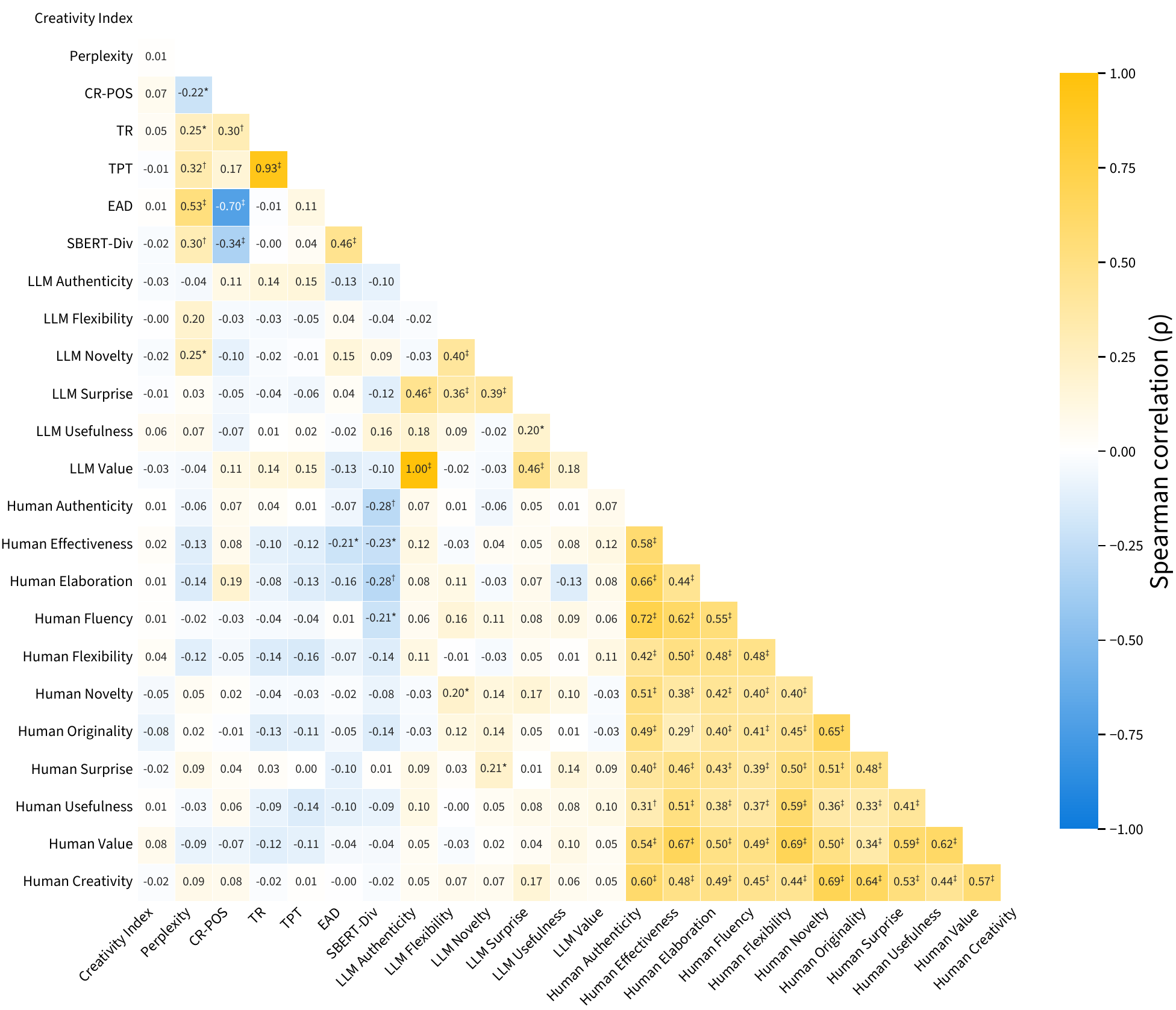 }
    \caption{\textbf{Extended correlation matrix for LLM-generated stories.} Spearman correlation ($\rho$) between all automatically computed evaluation metrics and subjective dimensions evaluated exclusively on the subset of texts generated by Large Language Models. Metrics exhibiting zero variance within this subset (such as constant maximum scores given by the LLM-as-a-Judge) were excluded from the analysis. Statistical significance is denoted by single-character typography: * ($p < .05$), $\dagger$ ($p < .01$), and $\ddagger$ ($p < .001$).}
    \label{fig:full_correlation_matrix_llm}
\end{figure}

\begin{figure}[htbp]
    \centering
    \includegraphics[width=1\textwidth]{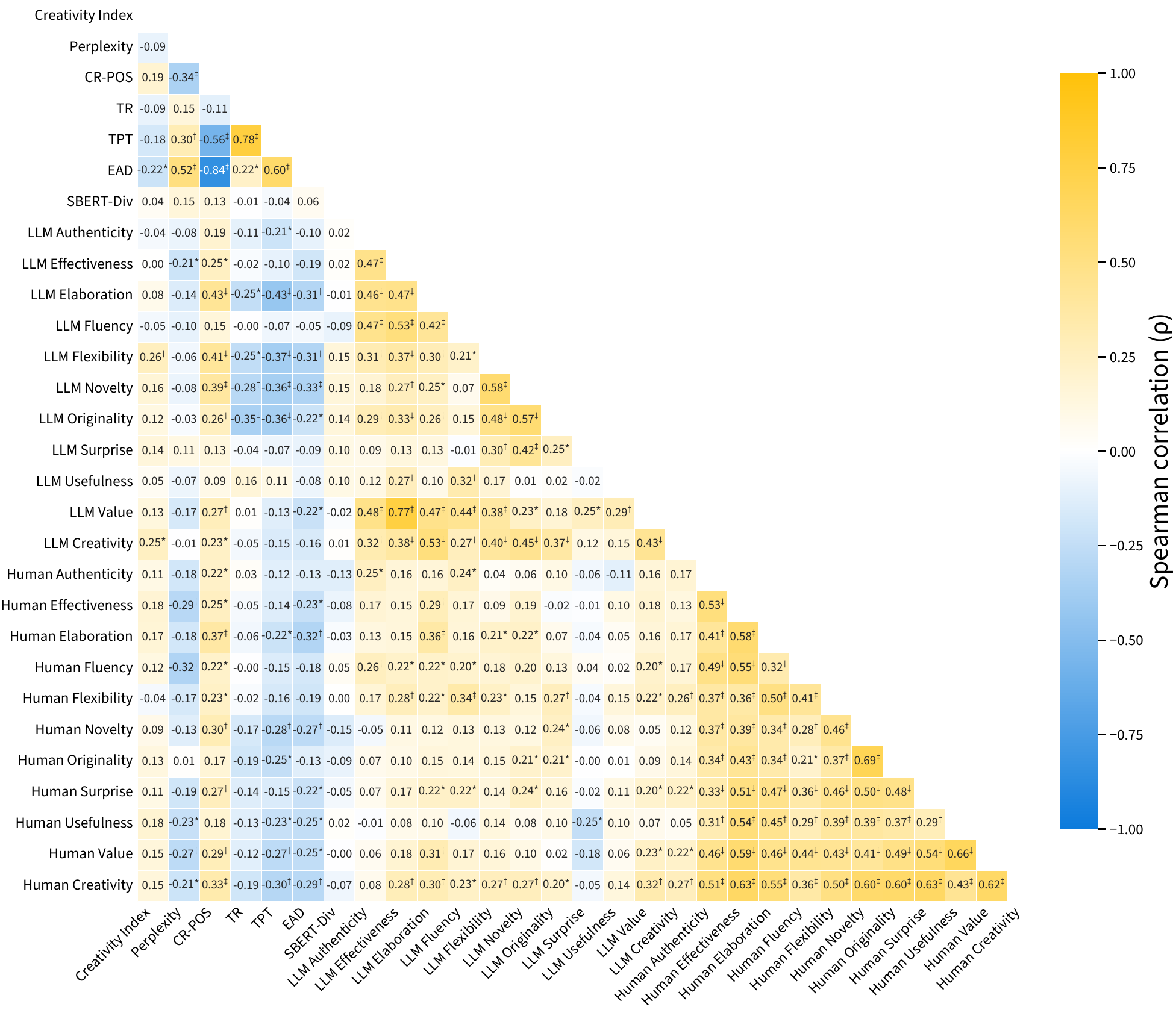 }
    \caption{\textbf{Extended correlation matrix for human-authored stories.} Spearman correlation ($\rho$) between all automatically computed evaluation metrics and subjective dimensions evaluated exclusively on the subset of texts written by humans. Statistical significance is denoted by single-character typography: * ($p < .05$), $\dagger$ ($p < .01$), and $\ddagger$ ($p < .001$).}
    \label{fig:full_correlation_matrix_human}
\end{figure}

In addition to the overall correlation analysis presented in the Results section, we provide the full correlation matrices stratified by text provenance. Figure \ref{fig:full_correlation_matrix_llm} shows the cross-correlation between subjective and objective metrics for LLM-generated stories only. In contrast, Figure \ref{fig:full_correlation_matrix_human} presents the same analysis for human-authored stories.

While the general trend is as discussed in the main body, with no machine-based metric showing a strong correlation with human judgments, a comparison between Figures \ref{fig:full_correlation_matrix_llm} and \ref{fig:full_correlation_matrix_human} reveals additional insights. In particular, for LLM-generated stories, only SBERT-Div shows a correlation higher than $0.2$ in absolute value, exhibiting a weak but consistent negative correlation with human scores for authenticity, effectiveness, elaboration, and fluency. Similarly, only perplexity shows a very weak positive correlation with LLM-as-a-Judge scores. All other correlations are negligible, further demonstrating the limitations of automatic methods for assessing AI creativity.

On the contrary, while still lacking strong correlations, the results for the human-authored stories are more interesting. Only the Creativity Index and SBERT-Div show very weak or no correlation with human scores; TR shows no correlation with the most qualitative dimensions and a weak negative correlation with the others, especially originality, creativity, and novelty. In contrast, CR-POS shows a positive correlation with all dimensions, peaking at elaboration and creativity, while Perplexity, TPT, and EAD show negative correlations with almost all dimensions, peaking at fluency and effectiveness, novelty and creativity, and elaboration and creativity, respectively. Similar results are also apparent when comparing automatically computed evaluation metrics with LLM-as-a-Judge scores.

\newpage
\section{Human survey statistics} \label{app:human_survey}

As reported in the Methods section, human evaluations were gathered via a custom-built web interface. In total, we collected 441 individual evaluations from 115 unique participants. To ensure statistical robustness, these evaluations were distributed across the entire dataset, resulting in an average of 2.21 independent human ratings per text. Before participation, we collected some background information to assess the reliability and diversity of our human baseline, which was stored in a fully anonymized manner. The demographic and linguistic profile of the participants is summarized in Table \ref{tab:demographics}. The collected demographics indicate that our evaluator pool has a robust educational background (with nearly 65\% holding a university degree) and a heterogeneous degree of familiarity with LLMs. This distribution ensures that the subjective ratings reflect a highly balanced mix of expert and general-user perspectives.

\begin{table}[htbp]
    \centering
    \renewcommand{\arraystretch}{1.1}
    \begin{tabular}{llcc}
        \toprule
        \textbf{Demographic Feature} & \textbf{Category} & \textbf{Count ($N$)} & \textbf{Percentage (\%)} \\
        \midrule
        \multirow{2}{*}{\textbf{Gender}} 
        & Male & 68 & 59.1\% \\
        & Female & 47 & 40.9\% \\
        \midrule
        \multirow{4}{*}{\textbf{Age}} 
        & 18--24 & 31 & 27.0\% \\
        & 25--34 & 24 & 20.9\% \\
        & 35--44 & 7 & 6.1\% \\
        & 45+ & 53 & 46.1\% \\
        \midrule
        \multirow{6}{*}{\textbf{Education Level}} 
        & Less than High School & 10 & 8.7\% \\
        & High School & 30 & 26.1\% \\
        & Bachelor's Degree & 21 & 18.3\% \\
        & Master's Degree & 49 & 42.6\% \\
        & PhD & 3 & 2.6\% \\
        & Other & 2 & 1.7\% \\
        \midrule
        \multirow{3}{*}{\textbf{Native Language}} 
        & Italian & 105 & 91.3\% \\
        & English & 8 & 7.0\% \\
        & Other & 2 & 1.7\% \\
        \midrule
        \multirow{5}{*}{\textbf{LLM Experience}} 
        & Never used & 25 & 21.7\% \\
        & Tried out of curiosity & 8 & 7.0\% \\
        & Occasionally & 36 & 31.3\% \\
        & Often & 26 & 22.6\% \\
        & If not daily, almost & 20 & 17.4\% \\
        \bottomrule
    \end{tabular}
    \caption{\textbf{Statistics of human evaluators.} Aggregated summary on the demographic, education, linguistic, and LLM familiarity background of the 115 human evaluators.}
    \label{tab:demographics}
\end{table}

\newpage

\section{LLM generation prompt}\label{app:prompts_gen}
To ensure reproducibility of the creative writing task, the following prompt was employed to generate the dataset of AI-authored stories across all five state-of-the-art models used, i.e., \texttt{GPT-5.2}, \texttt{DeepSeek-V3.2}, \texttt{Mistral Large 3}, \texttt{Claude Sonnet 4.5}, and \texttt{Gemini 3 Pro}. Specific constraints on formatting (Markdown) and length variation were introduced to prevent the models from producing texts of nearly identical length across generations (a common pitfall of repeatedly querying the same base prompt).

\begin{tcolorbox}[colback=ourblue!10!white,colframe=ourblue,title=System Prompt and Instructions for Creative Writing Task]
Act as a creative writer. I will provide you with a story premise. You must write a narrative text based on that premise adhering strictly to the following constraints:\\
\\
1. **Format:** The output must be in Markdown. I need to be able to copy the markdown. I want to see "\textbackslash n" at the end of the paragraph. Do not use "", use '' instead.\\
2. **Paragraphs:** Use standard spacing (new lines) to indicate new paragraphs.\\
3. **Length Constraint:** The text must be between 200 and 500 words.\\
4. **Randomness Factor:** Do not default to the minimum (200) or maximum (500) word count. You must simulate a random length within that range (e.g., 235, 435, 323) so the story ends naturally without feeling forced or padded.\\
\\
Here is the story premise:\\
\{premise\}
\end{tcolorbox}

%
%

\newpage
\section{LLM-as-a-Judge system instructions}
\label{app:prompts_judge}

The following interaction template was used to enforce the isolated evaluation strategy described in the Methods section. It consists of a fixed system prompt and a dynamic user prompt containing the definition of the specific metric and the input text.

\begin{tcolorbox}[enhanced jigsaw,breakable,colback=ourblue!10!white,colframe=ourblue,title=Full Interaction Template for LLM-as-a-Judge Evaluation]
\texttt{system}\\
You are an objective text evaluator. Your task is to assess a given text objectively according to a specific metric.\\
Your only output must be a single, syntactically correct JSON object.\\
Never include explanations, reasoning, or extra text outside the JSON.\\
\texttt{user}\\
Given the following text, you need to evaluate it for the specified metric: \{metric\}.\\
\\
\verb|#|\verb|#|\verb|#| Metric Definition\\
**\{metric\_name\}**: \{definition\}\\
\\
\verb|#|\verb|#|\verb|#| Task\\
Taking into account the definition above, produce:\\
- score: integer $1-5$ ($1$ = lowest, $5$ = highest)\\
- justification: $\leq30$ words explaining the rating\\
- excerpt: $\leq20$ words from the text supporting the evaluation\\
\\
\verb|#|\verb|#|\verb|#| Rules\\
Strict rules:\\
• Evaluate the aspect objectively based ONLY on the provided definition.\\
• Do NOT reveal chain-of-thought.\\
• If the text is ambiguous or too short to be judged, score 3 and note "insufficient evidence".\\
• Return **only valid JSON** with field: "name of the metric". The field must be an object with keys: score (int), justification (string), excerpt (string or null).\\
• Do NOT answer anything else other than the JSON.\\
• Do NOT include backticks, markdown, explanations, or anything outside the JSON.\\
• If unsure about JSON syntax, default to minimal valid JSON with null excerpt.\\
• Before answering, double check the brackets and the correctness of the JSON\\
• CRITICAL JSON RULE: All JSON keys and string values MUST be enclosed in double quotes (e.g., "justification": "your text here").\\
• CRITICAL TEXT RULE: Inside your justification or excerpt, do NOT use double quotes. If you need to quote a character, use single quotes (e.g., "excerpt": "He said 'hello'").\\
\\
Output structure:\\
\{\{\\
\hspace*{.5cm} "\{metric\}": \{\{\\
\hspace*{1.cm} "score": \textless int\textgreater,\\ 
\hspace*{1.cm} "justification": "\textless string\textgreater",\\
\hspace*{1.cm} "excerpt": "\textless string or null\textgreater"\\
\hspace*{.5cm} \}\}\\
\}\}\\
\\
SCALE ANCHORS (use these as guidance):\\
• $5$ = clear, strong, unambiguous evidence for the aspect.\\
• $4$ = good evidence, minor weaknesses.\\
• $3$ = ambiguous or mixed evidence; could go either way.\\
• $2$ = weak evidence or some counter-evidence.\\
• $1$ = no evidence or direct counter-evidence.\\
\\        
INPUT:\\
Text to evaluate:\\
"\{text\}"\\
\\
OUTPUT:\\
- JSON object (as described).\\
- The JSON must start with '\_curly bracket\_' and end with exactly one '\_curly bracket\_'.\\
- Ensure the correctness of the JSON. Double check that the brackets are correct.\\
\\
End.\\
\texttt{assistant}
\end{tcolorbox}

\newpage

\end{appendices}

\bibliography{sn-bibliography}

\end{document}